\documentclass[journal]{IEEEtran}
\usepackage{orcidlink}
\usepackage{cuted} 
\usepackage{subcaption}
\usepackage{cite}
\usepackage{amsmath}
\usepackage{graphicx}
\usepackage{url}
\usepackage{booktabs}
\usepackage{xcolor}
\usepackage[normalem]{ulem}
\usepackage[compatibility=false]{caption} 
\usepackage{xurl}
\begin{document}

\title{From Articles to Publishers: Aggregating Language Model Predictions for News Source Reliability Inference}

\author{John~Bianchi\,\orcidlink{0009-0006-2582-1480},
        Manuel~Pratelli\,\orcidlink{0000-0002-9978-791X},
        Fabio~Pinelli\,\orcidlink{0000-0003-1058-6917},
        and Marinella~Petrocchi\,\orcidlink{0000-0003-0591-877X}%
\thanks{J. Bianchi is with the IMT School for Advanced Studies Lucca, 55100 Lucca, Italy (e-mail: john.bianchi@imtlucca.it).}%
\thanks{M. Pratelli is with the Institute of Informatics and Telematics, National Research Council (IIT-CNR), 56124 Pisa, Italy (e-mail: manuel.pratelli@iit.cnr.it).}%
\thanks{F. Pinelli is with the IMT School for Advanced Studies Lucca, 55100 Lucca, Italy (e-mail: fabio.pinelli@imtlucca.it).}%
\thanks{M. Petrocchi is with the Institute of Informatics and Telematics, National Research Council (IIT-CNR), 56124 Pisa, Italy, and with the IMT School for Advanced Studies Lucca, 55100 Lucca, Italy (e-mail: marinella.petrocchi@iit.cnr.it).}%
\thanks{Corresponding author: John Bianchi.}%
\thanks{This work is partially supported by IIT-CNR internal project SIMULARE `SIMULAzione delle Reazioni alla disinformazionE con LLM'.}%
}

\markboth{IEEE Transactions on Computational Social Systems,~Vol.~XX, No.~X, XXXX~2026}%
{Bianchi \MakeLowercase{\textit{et al.}}: Aggregating Language Model Predictions for News Source Reliability Inference}

\maketitle

\begin{abstract}
Traditionally, the reliability of news publishers is assessed by expert organisations that evaluate editorial practices, transparency and factual standards at source. When this process is translated into a computational approach, the problem is often formulated at the level of individual articles, with models being trained on a set of pre-labelled articles and their performance being evaluated in a test phase. 
In this work, we investigate news source reliability inference as a source-level prediction problem. We propose a two-stage framework in which transformer-based language models first estimate the reliability of individual articles and subsequently aggregate article-level predictions to infer the reliability of previously unseen publishers.
To approximate realistic deployment conditions, we enforce a strict publisher-disjoint evaluation protocol, ensuring that no publisher appears in both training and test sets.
Experiments on 19,476 political news articles from 439 English-language publishers labeled with NewsGuard reliability ratings show that aggregation substantially improves robustness and performance, increasing accuracy from approximately 0.60 at the article level to 0.69 at the publisher level.
Finally, we analyze how prediction errors vary across political orientations, revealing statistically significant associations between political leaning and misclassification patterns.
Overall, our findings show that publisher reliability can be inferred from aggregated textual signals alone, supporting scalable and content-based approaches to automated news source assessment.

\end{abstract}


\IEEEpeerreviewmaketitle

\section{Introduction}
In contemporary digital media ecosystems, assessing the reliability of news publishers has become increasingly important for limiting the spread of misleading or low-quality information. Professional organizations such as NewsGuard, the Global Disinformation Index, and Media Bias Fact Check evaluate news outlets according to journalistic standards including transparency, editorial accountability, factual accuracy, and correction practices. Despite methodological differences, these systems exhibit substantial convergence, providing a stable foundation for reliability labeling~\cite{lin2023high,DBLP:conf/icwsm/PratelliP22}. Importantly, these assessments are performed at the level of the publisher rather than at the level of individual articles.

A large body of computational approaches to misinformation and credibility assessment formulates the problem as an article-level classification task~\cite{pandolfi2020automatic,Przybyla2021plugin,DBLP:journals/corr/abs-2410-21360,DBLP:journals/dbsk/AdjiLS26}. Existing systems typically attempt to determine whether a single article is credible, misleading, or unreliable based on linguistic patterns, stylistic signals, or neural representations. While effective in controlled settings, this formulation only partially reflects the actual structure of professional reliability assessment, where reliability is considered a persistent property of the news source emerging across multiple publications rather than an intrinsic property of isolated articles.

This mismatch introduces an important methodological challenge. Individual articles provide only partial and noisy evidence about the overall reliability of the publisher that produced them. Even highly reliable outlets may occasionally publish lower-quality content, while unreliable publishers may produce articles that appear factually plausible in isolation. Consequently, estimating publisher reliability from single articles is inherently unstable. A more realistic formulation instead consists in inferring source-level reliability by aggregating evidence collected across multiple articles belonging to the same publisher.

A second limitation of prior work concerns evaluation methodology. Existing studies evaluate article reliability using datasets in which labels are derived from publisher-level annotations and where evaluation does not explicitly enforce publisher-disjoint splits, see, e.g., ~\cite{burdisso-etal-2024-reliability, DBLP:journals/dbsk/AdjiLS26,batDataset}. 
Under these conditions, models can partially memorize publisher-specific patterns rather than learning transferable indicators of reliability~\cite{azizov-etal-2024-safari,verhoeven2025yesterday}.

In this work, we address these limitations by reformulating news reliability assessment as a publisher-level inference problem. We propose a two-stage framework in which transformer-based language models first estimate the reliability level of individual articles and subsequently aggregate article-level predictions to infer the reliability of previously unseen publishers. To approximate realistic deployment conditions, we enforce a strict publisher-disjoint evaluation protocol, ensuring that no publisher appears in both training and test sets.

More specifically, we frame article reliability estimation as a five-level ordinal classification problem aligned with NewsGuard ratings and evaluate four transformer-based architectures under strict cross-publisher generalization. We then study how aggregating predictions across collections of articles improves the estimation of publisher reliability. Finally, we analyze the relationship between political orientation and model errors, providing a bias-aware evaluation of the proposed framework.

Our contributions are fourfold:

\begin{enumerate}
    \item 
We reformulate news reliability assessment as a source-level inference problem, where publisher reliability is estimated by aggregating predictions from multiple articles.
\item We introduce a strict publisher-disjoint evaluation protocol that prevents source leakage and measures generalization to previously unseen publishers.
\item We construct a large-scale dataset of 19,476 political news articles from 439 English-language publishers labeled with NewsGuard reliability scores.
\item We show that aggregating article-level predictions substantially improves reliability estimation, achieving up to 0.69 publisher-level accuracy on unseen sources.\end{enumerate}

\section{Related Work}

The quality of online news has been widely studied in computational social science and natural language processing. Early work focused on article-level classification, detecting misleading content through linguistic features, supervised models, or deep neural architectures~\cite{pandolfi2020automatic}, often without explicitly considering the role of the publisher.

\paragraph{Publisher Reliability as a Core Dimension} A parallel line of research shifts the focus to the reliability of news publishers. This concept is described using terms such as \emph{reliability}, \emph{trustworthiness}, and \emph{credibility}, which capture whether a source adheres to journalistic standards and is perceived as informative rather than misleading~\cite{digitalmedia2007,infodisorder2017,10.1145/302979.303001}. Empirical studies show that source reputation strongly influences perceived information quality~\cite{doi:10.1080/07421222.2019.1628921}. 

\paragraph{Different signals to estimate publisher reliability} Computational approaches model publisher reliability using different types of signals. Some rely on textual features and stylistic patterns~\cite{przybyla2021classification}, while others incorporate external information such as user interactions, social networks, and other metadata~\cite{bazmi2023multi,DBLP:conf/ecir/PratelliSP25,DBLP:conf/icwsm/WilliamsCC25,DBLP:journals/jdiq/LongoZCZC25}. In addition, journalistic rating systems such as NewsGuard, the Global Disinformation Index, and Media Bias/Fact Check show strong agreement and are commonly used as ground truth for automated models~\cite{lin2023high,DBLP:conf/icwsm/PratelliP22}. Recent work further explores textual and LLM-based approaches to reliability estimation~\cite{10.1145/3770077,nakov2024survey,yang2023large,Leite2025}.

\paragraph{Datasets and Cross-Domain Generalization} Datasets with explicit publisher reliability labels remain limited. Bohacek et al.~\cite{bohavcek2022fine} introduce a Czech dataset with four reliability levels, while Adji et al.~\cite{DBLP:journals/dbsk/AdjiLS26} reconstruct the BAT corpus~\cite{batDataset} by recovering article texts and consolidating reliability categories into a standardized benchmark. More recent work highlights the challenge of generalization across domains and publishers~\cite{azizov-etal-2024-safari,burdisso-etal-2024-reliability,verhoeven2025yesterday}, showing that model performance degrades under distribution shifts.

Existing approaches may evaluate models in settings where training and test data share the same publishers. As a result, models may implicitly learn publisher-specific patterns rather than generalizable signals, leaving open whether reliability can be inferred from article text alone and transferred to unseen sources.

\paragraph{Our contribution} In contrast, we focus exclusively on textual content and enforce a strict publisher-disjoint evaluation protocol. We show that publisher reliability can be inferred from article text through aggregation of article-level predictions under a strict publisher-disjoint protocol.

\section{Problem Definition and Methodology}
\label{sec:problemDef}

\subsection{Problem definition}\label{subsec:problemdef}
We formulate the problem as `News Source Reliability Inference', whose objective is to estimate the reliability level of a publisher from its published content. Because publishers are observed through collections of articles, we decompose the problem into two stages: (i) Article-level reliability estimation and (ii) Publisher-level reliability inference through aggregation.

Let \(A = \{a_1, a_2, \ldots, a_n\}\) be a corpus of news articles. Each article \(a \in A\) has a textual representation \(t_a \in \mathcal{T}\) and belongs to a publisher \(p(a) \in P\). 
For a given publisher \(p \in P\), we denote by
\[
A_p = \{\, a \in A \mid p(a) = p \,\}
\]
the set of articles published by \(p\).
Each publisher \(p \in P\) has a fixed reliability level \(r_P(p)\). We use five ordered categories \(L = \{l_1, \ldots, l_5\}\) obtained by binning continuous reliability scores (see Section~\ref{sec:dataset}). Since reliability is a publisher's property, every article inherits its publisher's label: \(r_A(a) = r_P(p(a))\).
Analogously, each publisher \(p \in P\) has a political leaning associated \(o_P(p)\). 
The political leaning of each publisher is inherited by its articles \(a \in A_p\), formally: \(o_A(a) = o_P(p(a))\). The political leaning belongs to a set of 5 possible values \(O = \{\text{Far Left},\text{Slightly Left}, \text{Not Labeled}, \text{Slightly Right}, \text{Far Right} \}\). 
These two inheritance functions are needed since article-level annotation is not available. Notice that political leaning is not used as a model input or prediction target; rather, we retain this annotation to enable a post-hoc analysis of the classifier behavior (see Section~\ref{subsec:politicalBias}).

\paragraph{Stage 1: Article-level reliability estimation} 
We train a classifier \(C_{\text{reliability}}: \mathcal{T} \rightarrow L\) to predict the reliability level of a single article using only its text.

\paragraph{Stage 2: Publisher reliability inference} 

To simulate how a system profiles a new, unseen source, we estimate the publisher-level reliability by aggregating predictions obtained from its articles.
For a given publisher \(p \in P\), we aggregate predictions obtained from the articles in \(A_p\), defining an aggregation function
\[
Agg: L^{|A_{p}|} \rightarrow L
\]
that maps the article-level reliability predictions to a
publisher-level label.
The predicted reliability of publisher \(p\) is then computed as:
\[
\hat{r}_P(p) = Agg\big(\{ C_{\text{reliability}}(a) \mid a \in A_p \}\big).
\]
In~\ref{sec:results} we instantiate \(Agg\) as the mode, corresponding to majority voting. 

A key constraint of our problem formulation follows directly from the inheritance mechanism used to obtain article-level labels. Since each article inherits the reliability label of its publisher, articles originating from the same source are not independently labeled. Consequently, a random split at the article level would lead to severe label leakage, allowing the model to indirectly observe the test labels during training through articles from the same publisher. To address this issue, we enforce a strict separation of publishers between the training and test sets.
Let \(A_{\text{train}} \subset A\) and \(A_{\text{test}} \subset A\) denote
the sets of articles used for training and testing, respectively, with
\(A_{\text{train}} \cap A_{\text{test}} = \emptyset\).

Let \(P_{\text{train}} = \{\, p(a) \mid a \in A_{\text{train}} \,\}\) and
\(P_{\text{test}} = \{\, p(a) \mid a \in A_{\text{test}} \,\}\) be the sets
of publishers associated with the training and test articles.
We enforce that these sets are disjoint:
\begin{equation}\label{eq:publisher_disjoint}
P_{\text{train}} \cap P_{\text{test}} = \emptyset .
\end{equation}

This evaluation protocol ensures that all test articles originate from previously unseen publishers, thereby aligning the evaluation setting with the source-level nature of the ground-truth labels and preventing publisher-specific memorization.

\subsection{Evaluation Metrics and Baselines}\label{subsection:metrics}
We use the same metrics for both article-level and source-level evaluations.

\paragraph{Standard and Macro Metrics} 
We report Accuracy as the ratio of correct predictions. Although the class distribution is relatively stable, there are moderate differences in frequency (e.g., $l_4$ is approximately 2.6 times larger than $l_5$). To ensure that our evaluation remains robust against these disparities, we prefer Macro-averaged Precision, Recall, and F1-score. These metrics treat all five classes equally. Specifically, for each level $l$, we compute $\text{Recall}_l$ as the proportion of articles correctly assigned to level $l$ out of all articles truly belonging to that level. The Macro-Recall is then the arithmetic mean of these individual scores. This prevents the model from appearing successful by simply performing well on the most frequent categories at the expense of minority classes.

\paragraph{Tolerant Metrics} 

Given the ordinal nature of our labels, standard accuracy may be overly punitive for minor deviations. Therefore, we adopt \textit{Tolerant Accuracy}~\cite{wconf}, which assigns partial credit to predictions falling into adjacent categories. Formally, let $N$ be the total number of samples and $c_{ij}$ be the count of instances with true label $i$ predicted as $j$. Tolerant accuracy is defined as:

\begin{equation}
\mathrm{Accuracy}_{\mathrm{tolerant}}
= \frac{1}{N} \sum_{i \in L} \sum_{j \in L} w_{ij} \cdot c_{ij}
\label{eq:tolerant_accuracy}
\end{equation}

where the weight matrix $W = [w_{ij}]$ determines the credit assigned based on the distance between the true and predicted labels:

\begin{equation}
w_{ij} =
\begin{cases}
1 & \text{if } |i - j| = 0 \\
\alpha & \text{if } |i - j| = 1 \\
0 & \text{if } |i - j| > 1
\end{cases}
\end{equation}

Here, the parameter $\alpha \in [0, 1]$ controls the tolerance level. We report results for \(\alpha=0.5\) (half credit for adjacent errors) and \(\alpha=1.0\) (full credit, effectively treating adjacent classes as correct).

\paragraph{Baselines} 
To contextualize performance, we select baselines that establish strict lower bounds for `learning.' We pair specific baselines with the metrics they naturally challenge:

\begin{itemize}

    \item \textbf{Majority Baseline (for Accuracy):} This classifier assigns every instance to the most frequent class observed in the training set. It represents the toughest "zero-intelligence" competitor because it achieves the maximum possible accuracy without processing the input text, relying solely on label distribution. If a model fails to outperform this baseline, it indicates that the system has not learned any meaningful patterns from the text and is providing no information gain over a simple frequentist guess.
    
    \item \textbf{Uniform Random Baseline (for Macro Metrics):} This classifier assigns labels uniformly at random, giving each of the five reliability levels an equal probability of 20\%. We use this baseline for Macro metrics because Macro averaging treats all classes equally. A Majority classifier would be an unsuitable benchmark here: by always predicting the most frequent class, it would achieve zero Recall and F1-score for all other categories, resulting in a misleadingly low macro average. The Random baseline establishes a theoretical performance floor of $0.20$. Any model performance exceeding this threshold confirms that the system is capturing actual patterns in the text rather than relying on chance.
    
\end{itemize}

\subsection{Bias Assessment and Statistical Analysis}\label{subsec:bias_methodology}
Beyond aggregate performance, we investigate \textit{how} and \textit{why} models fail. We employ specific analytical frameworks to measure classification bias and its relationship with political orientation.

\paragraph{Error Analysis}

To detect structural weaknesses, we compute the Error Rate per reliability level $l \in L$, defined as $(1 - \mathrm{Recall}_l)$. Furthermore, to identify systematic over- or under-estimation, we analyze the Mean Error Direction (MED) for misclassified samples. Given a set of errors $E$, the MED is calculated as:
\begin{equation}
\mathrm{MED} = \frac{1}{|E|} \sum_{i \in E} (\hat{y}_i - y_i)
\end{equation}
where $\hat{y}_i$ and $y_i$ represent the predicted and true reliability levels (mapped to their ordinal values $1, \dots, 5$), respectively. Positive values indicate a tendency to overestimate reliability, while negative values indicate a tendency to underestimate reliability.

\paragraph{Statistical Significance of Political Bias}
To determine whether classification errors depend on the publisher's political leaning, we conduct a Pearson's $\chi^2$ test of independence. We test the null hypothesis ($H_0$) that error frequency is independent of political orientation. The statistic is computed as:
\begin{equation}
\chi^2 = \sum_{i,j} \frac{(O_{ij} - E_{ij})^2}{E_{ij}}
\end{equation}
where $O_{ij}$ is the observed frequency of outcome $j$ (Correct vs. Error) for political leaning $i$, and $E_{ij}$ is the expected frequency under independence.

To quantify the magnitude of this effect independent of sample size, we compute Cramér's V:
\begin{equation}
V = \sqrt{\frac{\chi^2}{N \cdot \min(r-1, c-1)}}
\end{equation}
where $N$ is the total number of samples, while $r$ and $c$ represent the number of rows (political orientations) and columns (classification outcomes) in the contingency table. Values of $V$ close to 0 imply no association, while values away from 0 indicate that political leaning strongly influences the likelihood of misclassification.

\section{Dataset and Labeling}
\label{sec:dataset}

This section describes the construction of the dataset used in our experiments. 
We detail the definition of the article set \(A\), the publisher set \(P\), and the reliability label set \(L\), together with the procedure used to assign reliability labels to articles.

\paragraph{Ground Truth: NewsGuard}
\label{sec:newsguard}

The publisher-level reliability function \(r_P : P \rightarrow L\) is derived from NewsGuard,\footnote{\url{https://www.newsguardtech.com/}} a professional media rating system in which trained journalists evaluate online news outlets.
Each publisher \(p \in P\) is assessed against nine widely adopted journalistic criteria, including the avoidance of deceptive headlines, the transparency of ownership and financing, and the adoption of effective correction practices.

The evaluation produces a continuous reliability score in the range \([0,100]\), which provides a quantitative measure of a publisher’s adherence to professional journalistic standards.\footnote{\url{https://www.newsguardtech.com/ratings/rating-process-criteria/}}  
The proprietary ratings were licensed for use to the authors of this publication.

\paragraph{Label Construction and Publishers Selection}
\label{sec:labelConstruction}

To obtain the discrete reliability label set
\(L = \{l_1, l_2, l_3, l_4, l_5\}\),
we discretize the continuous NewsGuard score into five ordinal categories, following the thresholds defined by NewsGuard. 
Table~\ref{tab:labels_domains} reports the mapping between score ranges and reliability labels, together with the number of publishers associated with each label after filtering for topic and language.

\begin{table}[ht]
\centering
\resizebox{\columnwidth}{!}{%
\begin{tabular}{llcccc}
\toprule
\textbf{Label} & \textbf{Score Range} & \textbf{Initial Dom.} & \textbf{Scraped Dom.} & \textbf{Final Art.} \\
\midrule
$l_1$ & 0--39 (Max Caution) & \sout{1,211} $\to$ 175 & 152 & 4,871 \\
$l_2$ & 40--59 (Caution)          & 127   & 69  & 3,016 \\
$l_3$ & 60--74 (Credible w/ Exc.) & 124   & 95  & 4,620 \\
$l_4$ & 75--99 (Generally Credible) & 175 & 88  & 5,041 \\
$l_5$ & 100 (High Credibility)    & 68    & 35  & 1,928 \\
\bottomrule
\end{tabular}%
}
\caption{Reliability label definition based on NewsGuard scores, including the number of domains after filtering for English-language political news outlets, the count of successfully scraped domains, and the total number of collected articles.}
\label{tab:labels_domains}
\end{table}

We restrict the publisher set \(P\) to outlets that (i) are tagged by NewsGuard as \emph{Political News and Commentary}, (ii) primarily publish content in English, and (iii)
do not enforce a paywall.
To prevent the article corpus from being dominated by highly unreliable sources, we cap the number of publishers in $l_1$ at 175, matching the number of publishers in $l_4$, the second most represented class.
After this balancing step, the final publisher set contains \(|P| = 669\) English-language political news outlets.

\paragraph{Article Collection}
\label{sec:dataCollection}

Let \(A\) denote the set of collected news articles.
For each publisher \(p \in P\), we scrape articles from its homepage using the \texttt{newspaper3k} library\footnote{\url{https://pypi.org/project/newspaper3k/}}.  
To ensure a balanced representation across publishers, we collect at most 200 articles per outlet.

The scraper is configured to retrieve English-language content only and to prevent duplicates. We rely on the automated article discovery features of the \texttt{newspaper3k} library to identify and extract content from the targeted domains\footnote{\url{https://newspaper.readthedocs.io/en/latest/}}.
As the extraction process follows the library's default traversal method, the order of retrieved articles does not necessarily correspond to a chronological timeline.
It is important to note that the scraping process was not successful for all targeted outlets. Some domains were inaccessible due to active anti-scraping measures (e.g., firewall blocks), server timeouts, or structural incompatibilities. Consequently, the number of distinct publishers in the final dataset is lower than the one in the original selection from NewsGuard.
To exclude non-news or low-information content, we retain only articles with a minimum length of 100 words, filtering out short announcements and advertisements. This procedure yields an initial corpus of 25,822 articles.

\paragraph{Data Curation and Labeling}
\label{sec:datacleaning}

To preserve the original linguistic characteristics of the articles, we apply no text normalization or cleaning.
A preliminary manual inspection of the collected data revealed the presence of non-English articles. This occurred despite explicitly configuring the scraper to request English content. The scraping process causes this inconsistency: the library requests the English version, but some domains erroneously return localized content or ignore the language header. The scraper then ingests the text without immediate validation.
We implement one preprocessing step to rectify this issue and ensure dataset consistency. We filter out non-English articles using the \texttt{langdetect} library\footnote{\url{https://pypi.org/project/langdetect/}}. This process removes 6,346 articles and leaves 19,476 English-language articles in the dataset.

Consistent with the problem formulation introduced in Section~\ref{subsec:problemdef}, each article \(a \in A\) is assigned a reliability label \(l_i \in L\) by inheriting the reliability level of its publisher, i.e., \(r_A(a) = r_P(p(a))\).

The final dataset composition is detailed in the last two columns of Table~\ref{tab:labels_domains}. While $l_1$ and $l_4$ originate from the same number of selected publishers (175 each), $l_1$ ultimately includes a larger number of active domains; $l_2$ and $l_3$ start from comparable numbers of publishers but differ substantially in the number of collected articles, whereas $l_5$ remains the least represented class, reflecting its limited presence in the original NewsGuard dataset.

\paragraph{Political Leaning Annotation}
\label{sec:politicalLeaningDataset}

In addition to reliability labels, NewsGuard provides a political leaning annotation for a subset of publishers.
Consistent with the problem formulation in
Section~\ref{subsec:problemdef}, this annotation is inherited at the article
level, i.e., for each article \(a\) published by \(p\), we set
\(o_A(a) = o_P(p(a))\).
Figure~\ref{fig:orientationDistributionInverted} shows the distribution of reliability labels across political leanings.
The pronounced imbalance highlights a structural property of the media ecosystem, whereby reliability levels are unevenly represented across political categories.
This characteristic must be considered when interpreting model performance and error patterns in subsequent analyses.

\section{Experiments}

This section presents the experimental setup and the results obtained
for the tasks defined in Section~\ref{subsec:problemdef}. In particular, we evaluate (i) \emph{Article-level reliability estimation} (Stage~1) under a
publisher-disjoint evaluation protocol, and (ii) \emph{Publisher reliability inference}
(Stage~2) obtained by aggregating article-level predictions. Model performance
is assessed using the evaluation metrics introduced in
Section~\ref{subsection:metrics}.

\subsection{Experimental Setup}
\label{sec:experimental_setup}

Following the validation protocol described in Section~\ref{sec:problemDef} (Equation~\ref{eq:publisher_disjoint}), we use a 5-fold cross-validation scheme with strict separation of publishers between training and test sets. This constraint directly follows from the inheritance mechanism used to obtain article-level labels: since all articles from the same publisher share the same reliability annotation, allowing articles from a publisher to appear in both training and test sets would lead to severe label leakage and publisher-specific memorization.
This evaluation setting assesses the model’s ability to generalize to articles from previously unseen publishers and approximates a realistic \textit{in-the-wild} deployment scenario, where the system encounters content
from sources not observed during training.
In addition, we employ a stratified cross-validation strategy using publisher reliability as the stratification variable, ensuring that each fold preserves the same reliability distribution and maintains identical proportions of publishers across reliability classes. To construct the folds, we retain all articles associated with each publisher. Although publishers are linked to varying numbers of articles, up to 200 per publisher, as described in Section \ref{sec:dataCollection}, we include them all to preserve signal diversity and avoid discarding potentially informative examples.
As a dispersion check, we compute, for each reliability class, the standard deviation of its proportion across the five folds. The resulting values are small (ranging from 1.37 to 3.61 percentage points), indicating that reliability class proportions are well preserved across folds.

We select four transformer-based models that embody different design principles in Natural Language Processing. This allows us to benchmark performance across standard encoder baselines, optimized encoders, and state-of-the-art decoder-based large language models:

\begin{itemize}
    \item \textbf{BERT (Bidirectional Encoder Representations from Transformers)}: 
    We adopt the BERT-base-uncased model \cite{devlin2019bert} as our primary baseline. As a quintessential Encoder-only architecture, BERT is designed to learn deep bidirectional context representations by attending to tokens on both the left and right simultaneously. This makes it naturally suited for discriminative tasks like text classification and establishes a robust lower bound for performance, ensuring comparability with a vast body of prior work \cite{rogers2021primer}.

    \item \textbf{RoBERTa (Robustly Optimized BERT Pretraining Approach)}: 
    We select RoBERTa-large to test the limits of the encoder-only paradigm. While retaining the bidirectional structure of BERT, it eliminates the Next Sentence Prediction (NSP) objective  \cite{liu2019roberta}. This model allows us to verify if optimized training strategies alone can improve reliability detection, even while keeping the underlying architecture unchanged.

    \item \textbf{Llama}: 
    We include the 8B version of Llama 3 \cite{grattafiori2024llama} to represent the shift toward decoder-only generative models. Unlike the discriminative nature of BERT and RoBERTa, Llama 3 is an autoregressive model. Our goal is to assess if the massive parameter scale of a general-purpose LLM translates into superior classification accuracy, comparing the efficiency of encoders against the extensive knowledge of generative decoders.

    \item \textbf{Mistral}: 
    We include Mistral 24B \cite{jiang2024mixtral} to evaluate the effect of further scaling within decoder-only architectures. With 24 billion parameters, roughly three times larger than Llama, Mistral allows us to test whether additional model capacity improves reliability detection or whether performance saturates given the characteristics of the data.

\end{itemize}

The four architectures described above instantiate the article-level classifier \(C_{\text{reliability}}: \mathcal{T} \rightarrow L\) defined in Section~\ref{subsec:problemdef}.

\paragraph{Training Strategies and Hyperparameters}
We employed distinct fine-tuning strategies tailored to the architectural differences between encoders and decoders:
\begin{itemize}
    \item \textbf{Encoders (BERT, RoBERTa):} We performed \textit{full fine-tuning}, updating all model parameters. We used the \texttt{BertTokenizerFast} and \texttt{RobertaTokenizerFast} respectively. Optimization was carried out using AdamW with a standard learning rate of $2e^{-5}$ and weight decay, consistent with established literature \cite{devlin2019bert, liu2019roberta}.

    \item \textbf{Decoders (Llama and Mistral):} Given the size of the models, performing full fine-tuning was computationally infeasible. Instead, we adopted Low-Rank Adaptation (LoRA), a Parameter-Efficient Fine-Tuning (PEFT) technique. We froze the pre-trained weights and only updated low-rank adaptation matrices injected into the attention layers. This allowed us to train with a higher learning rate of $2e^{-4}$ using mixed-precision (FP16) to further reduce the memory footprint. We utilized the causal \texttt{LlamaTokenizer} and \texttt{MistralTokenizer} with manual padding management. For Mistral we enabled Flash Attention 2 for efficiency.

\end{itemize}

Each model was trained for 5 epochs per fold, with the exception of Mistral, which was trained for 3 epochs. This duration was established following a preliminary convergence study (see Supplementary Material~\cite{bianchi2026supplementary}), which demonstrated that validation loss minimizes at Epoch 4 for Llama, while Mistral exhibits comparable rapid convergence within three epochs. Consequently, extending training beyond this point yields no performance gains and increases the risk of overfitting. To allow optimality, we implemented an early stopping mechanism that saves the model state with the lowest validation loss. Batch sizes were dynamically adjusted according to memory requirements: 32 (training) / 16 (eval) for BERT, 16/8 for RoBERTa, 8/8 for Llama (with gradient accumulation), and 6/48 for Mistral (with gradient accumulation steps=5). We enforced a fixed global random seed (42) to ensure identical data splits and deterministic training behaviors across all architectures. The maximum sequence length for input processing was set to 256 tokens, except for Mistral, for which it was increased to 1,024 tokens. This was done to retain substantially more article text and to test whether a longer context improves classification.

All fine-tuned models are publicly available to foster further research\footnote{\label{fn:model_availability}\url{https://osf.io/r9atz/overview?view_only=e4bda170a3e74ca3ae245475d4486d74}}.

\subsection{Stage 1: Article-level reliability estimation}
\label{sec:results}

We first evaluate the article-level estimators that constitute the first stage of the proposed framework. We report Accuracy, along with macro-averaged Precision, Recall, and F1-score, to account for class imbalance and ensure fair evaluation across classes. 
Furthermore, we consider a bespoke accuracy metric that gives partial credit to predictions in classes that are adjacent to the ground truth  (Section \ref{subsection:metrics}).

\begin{figure}[h!]
    \centering
    \includegraphics[width=1\columnwidth]{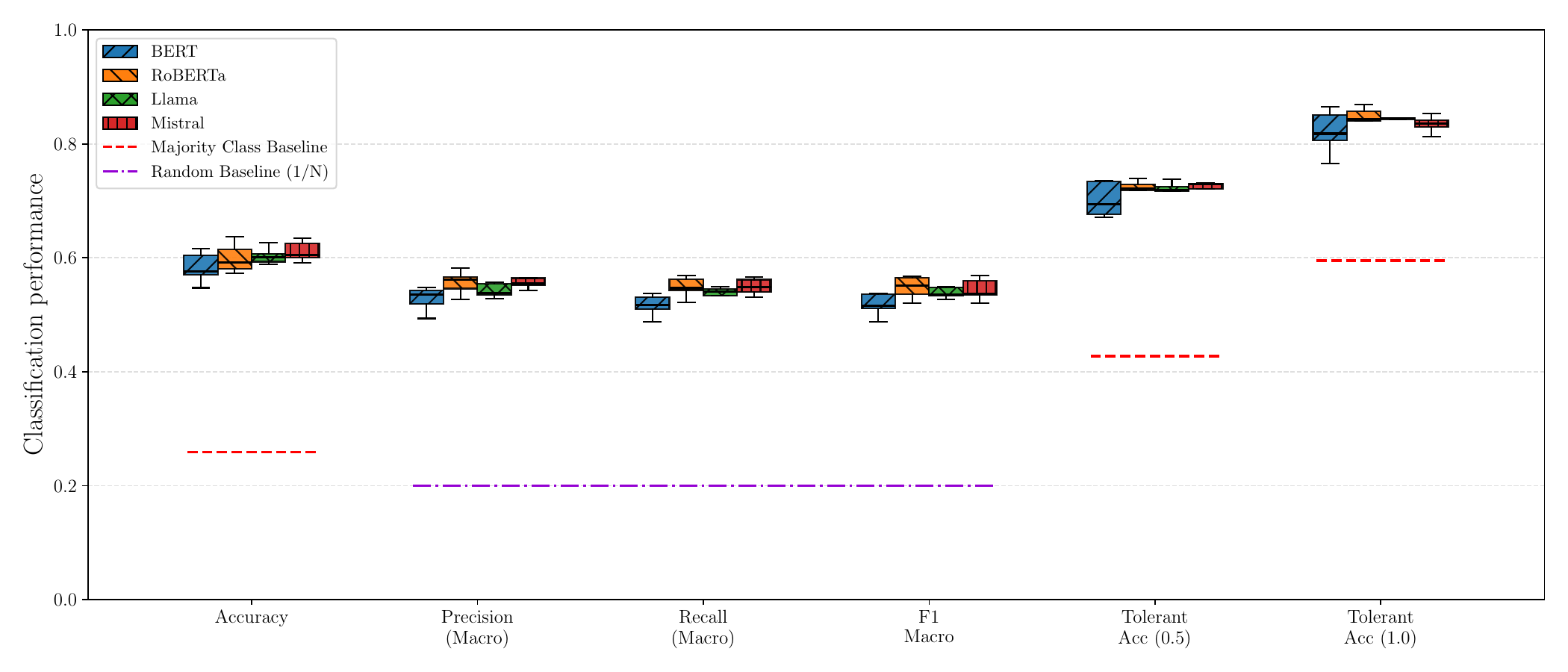}
    \caption{Distribution of evaluation metrics for BERT, RoBERTa, Llama, and Mistral across five folds. The boxplots report Accuracy, $\mathrm{Precision}_{\mathrm{macro}}$, $\mathrm{Recall}_{\mathrm{macro}}$, $\mathrm{F1}_{\mathrm{macro}}$, and two variants of $\mathrm{Accuracy}_{\mathrm{tolerant}}$ ($\alpha=0.5$ and $\alpha=1.0$). The horizontal lines represent the baselines: the dashed line is the Majority Class baseline (in red) and the dash-dotted line is the Random Baseline (in purple).}
\label{fig:results}
\end{figure}

Figure \ref{fig:results} illustrates the results. The dashed lines indicate the baselines. We distinguish between two references:
\begin{enumerate}
    \item \textbf{Majority Class Baseline (MCB)} (red): Used for Accuracy and Tolerant Accuracy. It represents a model that always predicts the most frequent label.
    \item \textbf{Random Baseline (RB)} (violet): Used for Macro metrics. Since macro-averaging treats all classes equally regardless of prevalence, the appropriate baseline is a uniform random guess ($1/N$).
\end{enumerate}
In terms of standard accuracy, all models achieve scores around 0.60, which is more than double the MCB. Mistral achieves the highest mean score of 0.61. Regarding macro-averaged performance, RoBERTa shows the highest mean $\mathrm{F1}_{\mathrm{macro}}$ of 0.55, followed by Mistral with 0.54. Llama and BERT rank just behind with 0.54 and 0.52, respectively. These scores significantly exceed the RB of 0.20. The variance across folds is minimal (standard deviation of approximately 0.02), indicating stability. The rightmost sections of the plot display tolerant accuracy. With a stricter penalty ($\alpha = 0.5$), Mistral, RoBERTa, and Llama reach approximately 0.72, which far exceeds the adjusted MCB of 0.43. 
When adjacent errors are considered fully correct ($\mathrm{Accuracy}_{\mathrm{tolerant}}$  = 1.0), RoBERTa achieves a top score of 0.84, followed by Mistral, Llama, and BERT with 0.83, 0.83, and 0.82, respectively. Although this metric's baseline is naturally higher (0.60), the models demonstrate significant performance gains (+0.24).
Results obtained for tolerant accuracy suggest that misclassifications are rarely severe, typically involving adjacent labels.

\subsection{Biases in Article-level reliability estimation}
\label{subsec:error_analysis}

Here, we study classification bias with respect to reliability level along two axes: (i) error frequency across levels, i.e., whether the classifier systematically performs better on some reliability classes than on others; and (ii) error direction, i.e., whether predictions tend to overestimate or underestimate the true reliability level.

\begin{figure}[h!]
    \centering
    \includegraphics[width=1\columnwidth]{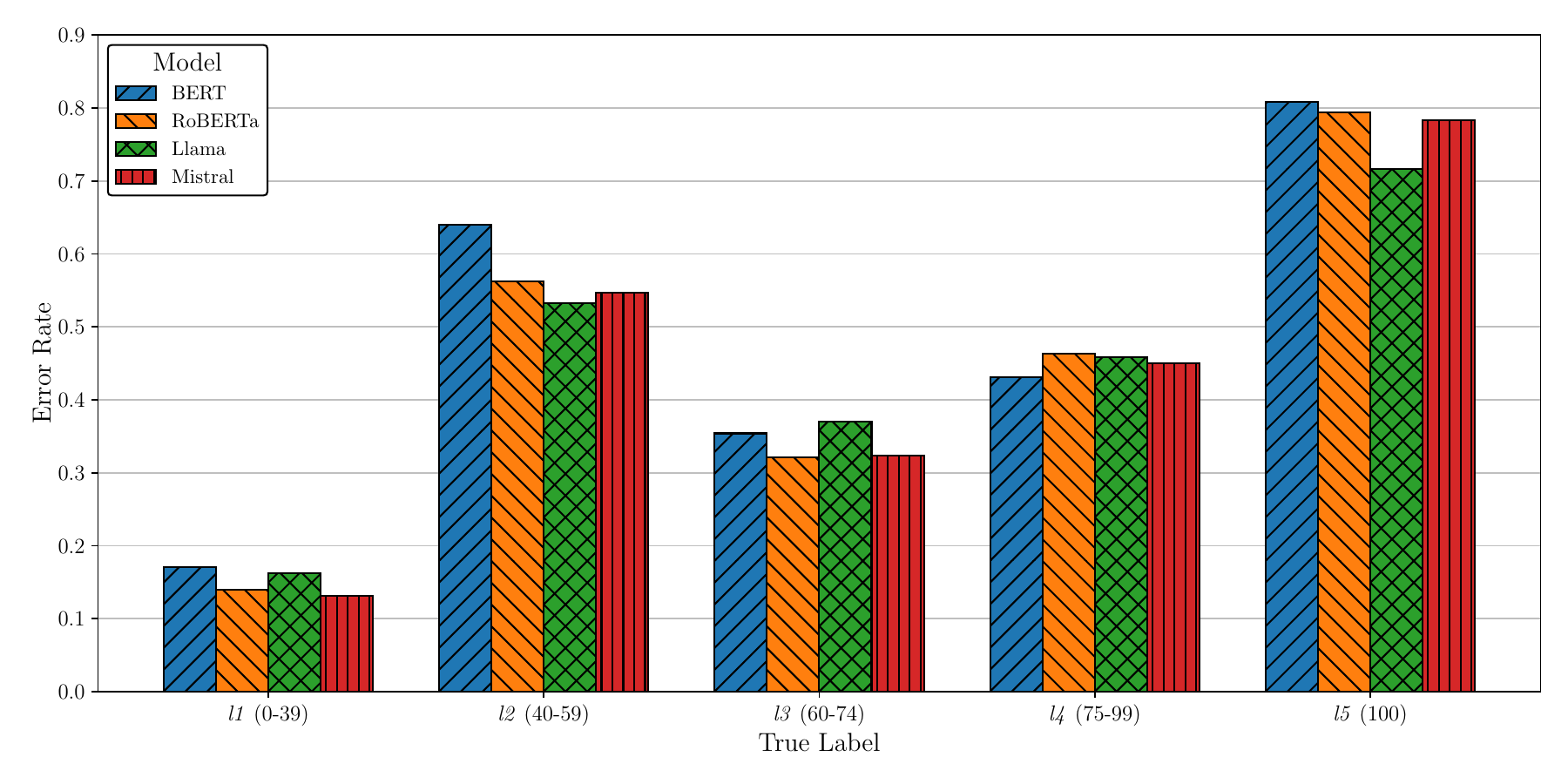}
    \caption{Per-class error rate comparison. The error rate is defined as $(1 - \mathrm{Recall}_l)$, where $\mathrm{Recall}_l$ is the proportion of articles correctly assigned to level $l$ out of all articles truly belonging to that level.}
    \label{fig:error_rate}
\end{figure}

Figure~\ref{fig:error_rate} highlights an asymmetry in prediction correctness. The lowest error rates are on $l_1$ (error rate $< 0.20$). This suggests that highly unreliable content has recognizable structural features.
The highest-reliability articles ($l_5$) are the most challenging to recognize for the model, with error rates exceeding 70\%\footnote{We remind the reader that class $l_5$ only includes articles that have been given a reliability score of 100. It could easily be confused with $l_4$, which contains articles with similarly high scores. This could motivate the high error rate.}. Class $l_2$ also proves difficult, with an error rate of over 50\%. Class $l_3$ sits in the middle, showing a moderate error rate of approximately 35\% and performing significantly better than classes $l_2$ and $l_5$.

Figure \ref{fig:error_direction} plots the Mean Error Direction (MED) to quantify the prediction bias. MED is defined as the average signed difference between predicted and true reliability labels over misclassified instances (see Section~\ref{subsec:bias_methodology}). 
We observe a tendency to regress towards the center. We attribute this to two structural factors:

\begin{figure}[h!]
    \centering
    \includegraphics[width=1\columnwidth]{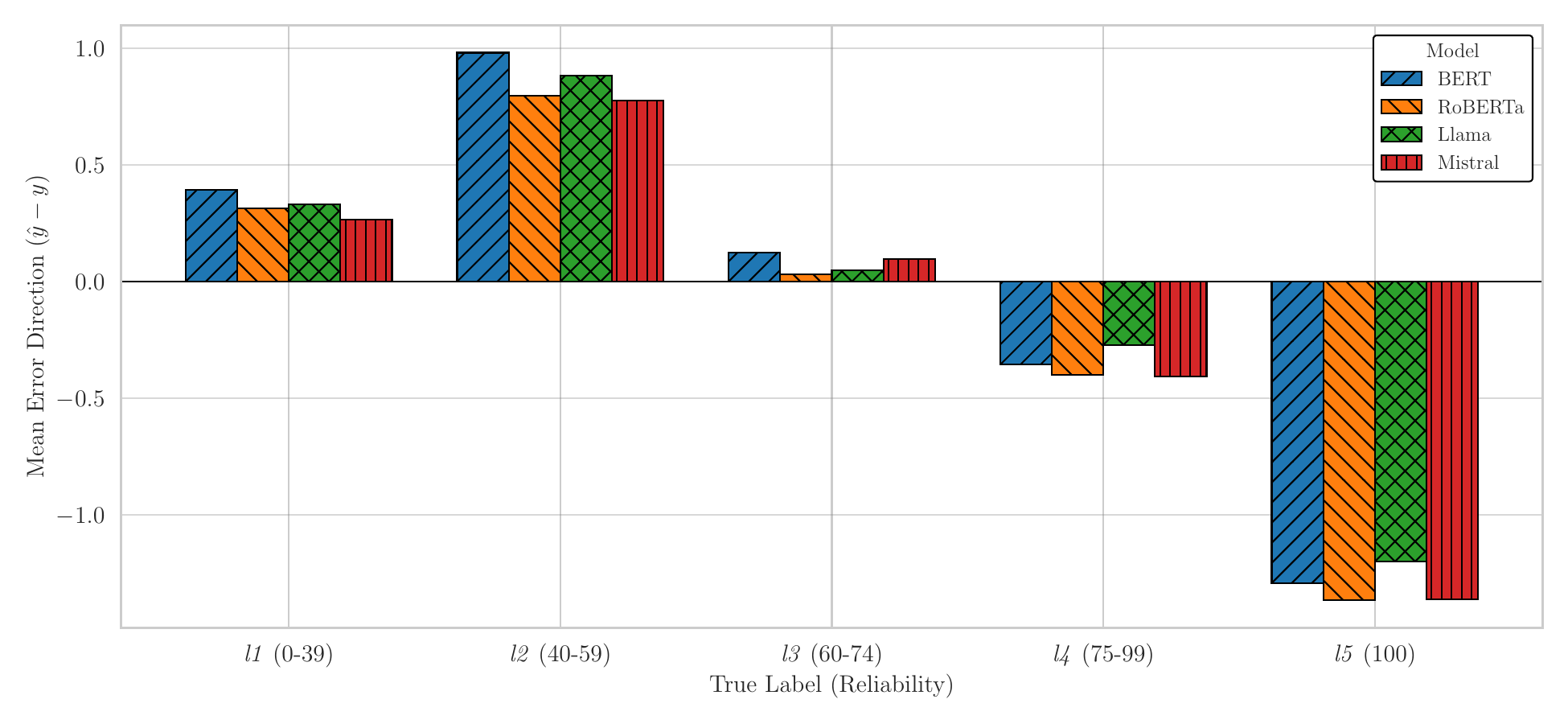}
    \caption{Mean Error Direction analysis. Positive values indicate overestimation of reliability, while negative values indicate underestimation.}    
    \label{fig:error_direction}
\end{figure}

\begin{itemize}
    \item \textbf{Boundary Constraints} ($l_1$ and $l_5$).
    Errors at the extremes are naturally constrained: models can only overestimate $l_1$ and underestimate $l_5$. However, the significant negative MED in $l_5$, $\approx -1.30 $ (BERT -1.29, RoBERTa -1.37, Llama -1.20, Mistral -1.36). The confusion matrices reveal that the vast majority of misclassifications for $l_5$ are concentrated in adjacent $l_4$ \cite{bianchi2026supplementary}. This confirms that the high error rate is primarily due to the models struggling to distinguish between perfect (100) and generally credible (75–99) reliability, likely because there is little textual distinction between these high-standard categories. The fact that the MED is slightly below -1.0 indicates a minor tendency to misclassify some $l_5$ instances into lower tiers (levels 2 and 3), but the main misclassification consists of labeling $l_5$  as $l_4$ . 
    
    \item \textbf{The `Safe Harbor' Effect ($l_3$).} 
    Class $l_3$ shows a Mean Error Direction close to zero (0.03 to 0.12 across models, with Mistral at 0.10). $l_3$ acts as a gravitational center for uncertainty: it absorbs a significant portion of misclassifications from the lower-reliability $l_2$ and from the higher-reliability $l_4$. Since the model overestimates $l_2$ (pushing it up to 3) and underestimates $l_4$ (pushing it down to 3), these opposing error vectors cancel each other out, resulting in a net directional bias of zero.    
\end{itemize}

These directional patterns generally remain consistent across all four models, including the substantially larger Mistral, suggesting that the biases are structural properties of the task rather than intrinsic to a specific architecture or model scale; the Supplementary Material~\cite{bianchi2026supplementary} reports full confusion matrices across all five cross-validation folds.

\subsection{Influence of Political Leaning on Classification Errors}\label{subsec:politicalBias}
\begin{figure}[h!]
    \centering
    \includegraphics[width=\columnwidth]{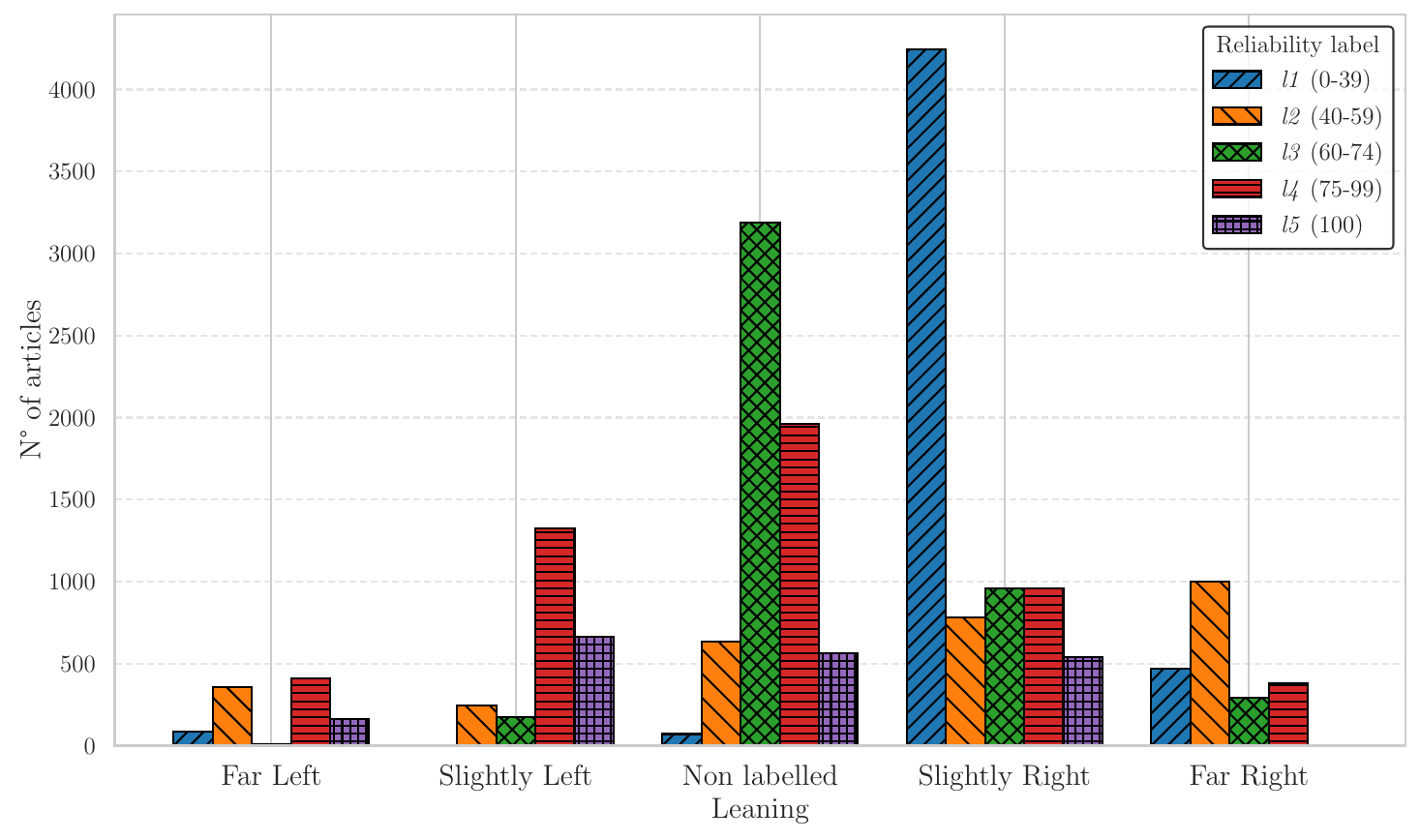}
    \caption{Distribution of reliability labels across political leanings. Each color and hatch pattern represents a distinct reliability level. Non labelled indicates articles from publishers not categorized by NewsGuard with respect to political leaning.}
    \label{fig:orientationDistributionInverted}
\end{figure}

The core objective of this section is to determine if the models exhibit political bias. Specifically, we investigate whether the models treat all sides of the political spectrum equally or if they penalize specific orientations.

\begin{figure}[t]
    \centering
    \includegraphics[width=\columnwidth]{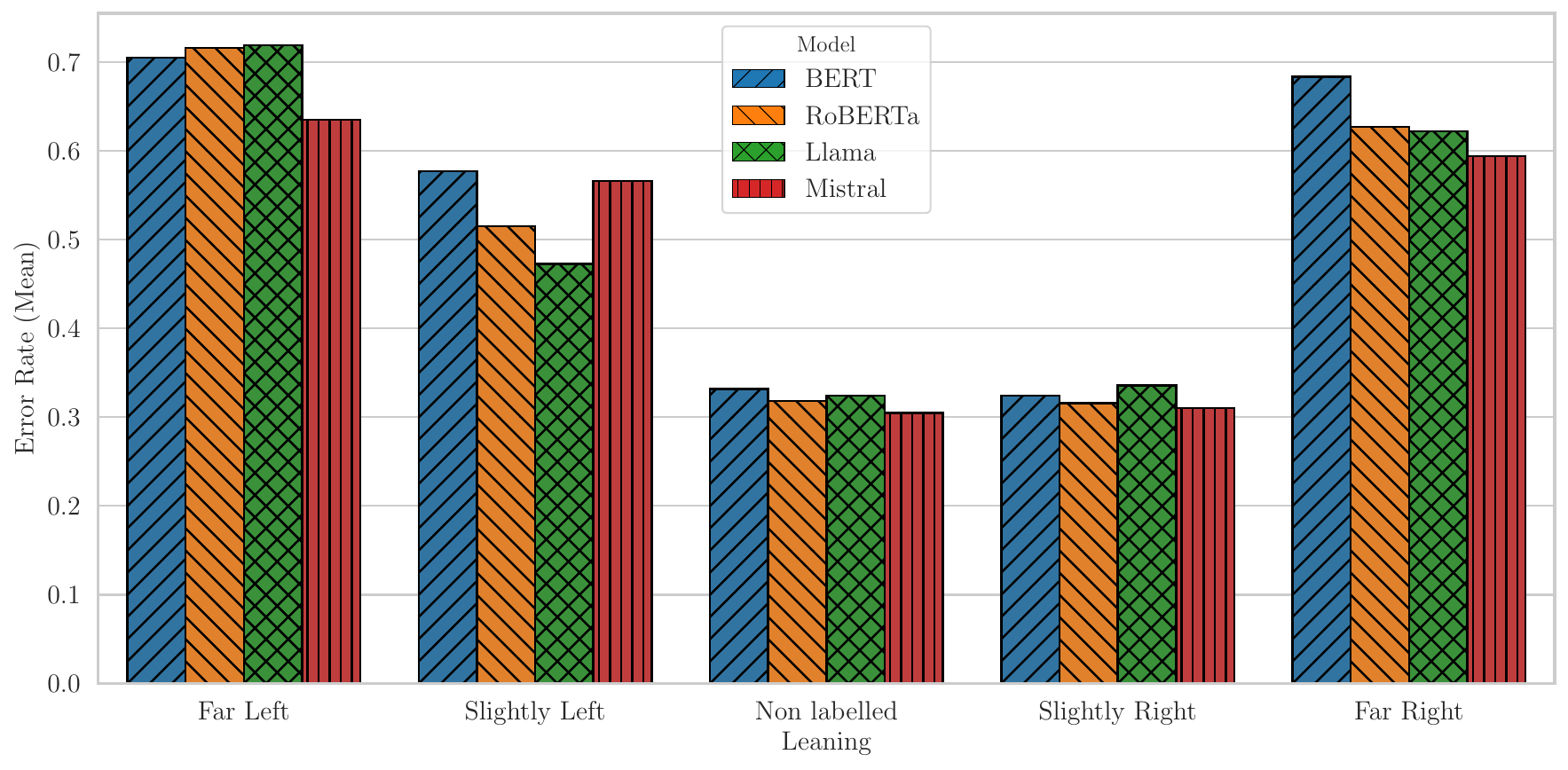}
    \caption {Error rate by political orientation and model. The bar chart displays the proportion of incorrect predictions for each model across the five political leanings. The error rate is calculated as the ratio of incorrect predictions to the total number of predictions (equivalent to the mean of a binary variable where 1 = incorrect and 0 = correct).}
    \label{fig:politicalErrorRates}
\end{figure}

\begin{figure}[t!]
    \centering
    \includegraphics[width=\columnwidth]{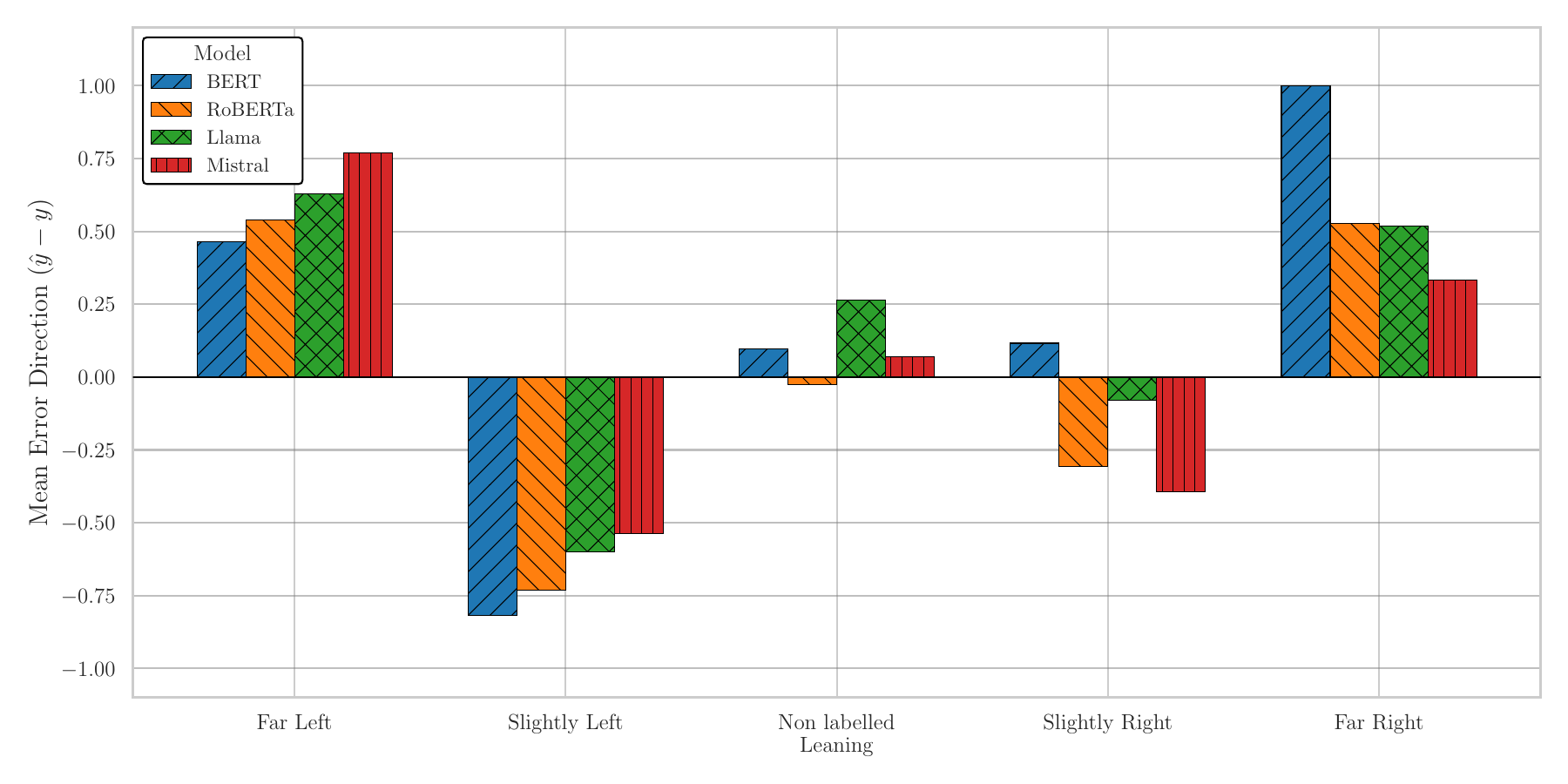}
    \caption{Mean value and direction of misclassifications, stratified by political leaning. Positive values indicate overestimation (prediction $>$ true label), while negative values indicate underestimation.}
    \label{fig:politicalErrorsDirection}
\end{figure}

We recall that, in the present study, the political orientation is a property of the publisher (i.e., the domain), not of the individual article.
Since the classifiers operate at the article level, the publisher’s political leaning is inherited by each article it produces, i.e., \(o_A(a) = o_P(p(a))\) (see Section \ref{subsec:problemdef}). This enables a post-hoc analysis of how source-level political alignment relates to article-level prediction errors.
Aimed at revealing whether the models behave uniformly across the political spectrum or whether notable political patterns emerge, this analysis proceeds in three steps:

\begin{enumerate}
    \item \textbf{Distribution Analysis.} First, we map how political orientations are distributed across reliability classes in our dataset (Figure \ref{fig:orientationDistributionInverted}).

    \item \textbf{Error Rate Analysis.} Second, we provide a quantitative assessment of the errors made by the models with respect to political orientation. Specifically, we evaluate whether misclassification errors are unevenly distributed across political leanings (Figure~\ref{fig:politicalErrorRates}) and test the statistical significance of these differences (Table~\ref{tab:chi_square_results}).

    \item \textbf{Directional Bias Analysis.} Finally, we quantitatively assess whether, among misclassified instances, the models exhibit systematic directional biases—specifically, a tendency to overestimate reliability (positive bias) or to underestimate it (negative bias)—as a function of political leaning (Figure~\ref{fig:politicalErrorsDirection}).
\end{enumerate}

\begin{table}[t]
\centering
\begin{tabular}{lcccc}
\textbf{Model} & \textbf{Accuracy} & $\boldsymbol{\chi^2}$ & \textbf{$p$-value} & \textbf{Cramér’s V} \\
\midrule
BERT & 58.27\% & 1684.13 & $<10^{-300}$ & 0.294 \\
RoBERTa & 60.34\% & 1418.31 & $<10^{-300}$ & 0.270 \\
Llama & 59.97\% & 1207.83 & $<10^{-250}$ & 0.249 \\
Mistral & 61.16\% & 1343.98 & $<10^{-300}$ & {0.263} \\
\end{tabular}
\caption{
Statistical tests for Political Leaning vs. Classification Correctness. The table reports the Chi-square ($\chi^2$) test of independence and the associated $p$-values to assess whether error rates vary across political orientations. Cramér’s V is provided to measure the effect size of this association.}
\label{tab:chi_square_results}
\end{table}

Figure~\ref{fig:orientationDistributionInverted} shows an imbalance across political orientations in the dataset. Accordingly, we compute error rates separately for each leaning, defined as the proportion of misclassified instances within each class. This class-conditional formulation makes error rates independent of class frequency, allowing us to analyze model errors despite the overall imbalance.

The resulting mean misclassification rates computed for each model and political orientation are reported in Figure~\ref{fig:politicalErrorRates}. We observe that all models exhibit systematic variation in performance across political leanings. There are substantially higher error rates for far left and far right outlets (59–72\%) compared to lower rates for slightly right and non-labeled outlets (30–34\%), and intermediate error rates for slightly left articles (47–58\%).  Mistral 24B deviates from the other models by showing fewer errors in the extreme categories and a higher error rate in the Slightly Left category. Its performance is comparable in the Non-labelled and Slightly Right categories.

To assess the statistical significance of the observed bias, we apply a chi-square test of independence. The null hypothesis assumes that classification correctness is independent of political leaning. For all models, this hypothesis is rejected (Table~\ref{tab:chi_square_results}), indicating that misclassification rates are significantly associated with political orientation. Cramér’s~$V$ further reveals a consistent small-to-moderate effect size.

As a third analysis, we focus on misclassified instances to assess whether models exhibit directional biases, namely systematic tendencies to overestimate or underestimate article reliability as a function of political leaning. Figure~\ref{fig:politicalErrorsDirection} reports the Mean Error Direction (MED), defined as the average signed difference between predicted and true reliability labels over misclassified instances for each orientation (see Section~\ref{subsec:bias_methodology}). As anticipated in Section \ref{subsec:error_analysis}, positive MED values indicate overestimation of reliability, whereas negative values indicate underestimation.

Clear directional patterns emerge across all four models. Far Left and Far Right articles exhibit consistent overestimation, with mean signed errors ranging from +0.33 to +1.00, approaching a full reliability level. Slightly Left articles show systematic underestimation, with average values between -0.54 and –0.82. Slightly Right and non-labeled outlets display smaller, less consistent deviations: Llama, RoBERTa, and Mistral slightly underestimate Slightly Right reliability (up to -0.39), whereas BERT slightly overestimates it; for non-labeled outlets, BERT, Llama, and Mistral exhibit mild positive biases, while RoBERTa's average bias is near zero; the Supplementary Material~\cite{bianchi2026supplementary} reports full confusion matrices stratified by political leaning.

\subsection{Stage 2: Publisher reliability inference}

The publisher-level inference task constitutes the primary objective of this work. The article-level estimators analyzed in the previous section are not the final target of the framework, but rather intermediate components used to infer the reliability of previously unseen news sources.

As anticipated in Section~\ref{sec:problemDef}, publisher-level reliability is obtained by aggregating article-level predictions for each publisher. To evaluate this approach, we use a simple aggregation function, \(Agg\), which is defined as the mode of a publisher’s article-level predictions (i.e., majority voting).
For a given publisher \(p \in P\), the system collects the predicted reliability labels \(C_{\text{reliability}}(a)\) for all articles \(a \in A_p\) in the test set (the predicted labels for the articles in each of the five test folds) and assigns the most frequent class as the final publisher-level label \(\hat{r}_P(p)\).
For example, if a publisher has ten articles and six are classified as \(l_4\) while four are classified as \(l_3\), the resulting publisher-level label is \(l_4\). This aggregation strategy is based on the hypothesis that combining predictions from multiple articles can offset the impact of variability or noise in individual classifications.

Our analysis considers all article-level predictions across the test sets, thereby simulating an \emph{in-the-wild} deployment scenario. In this scenario, the reliability of the publisher is inferred exclusively from the articles in the test set, ensuring that the model estimates the reliability of a source without having observed any of its content during training.

\begin{figure}[t]
  \centering
  \includegraphics[width=\columnwidth]{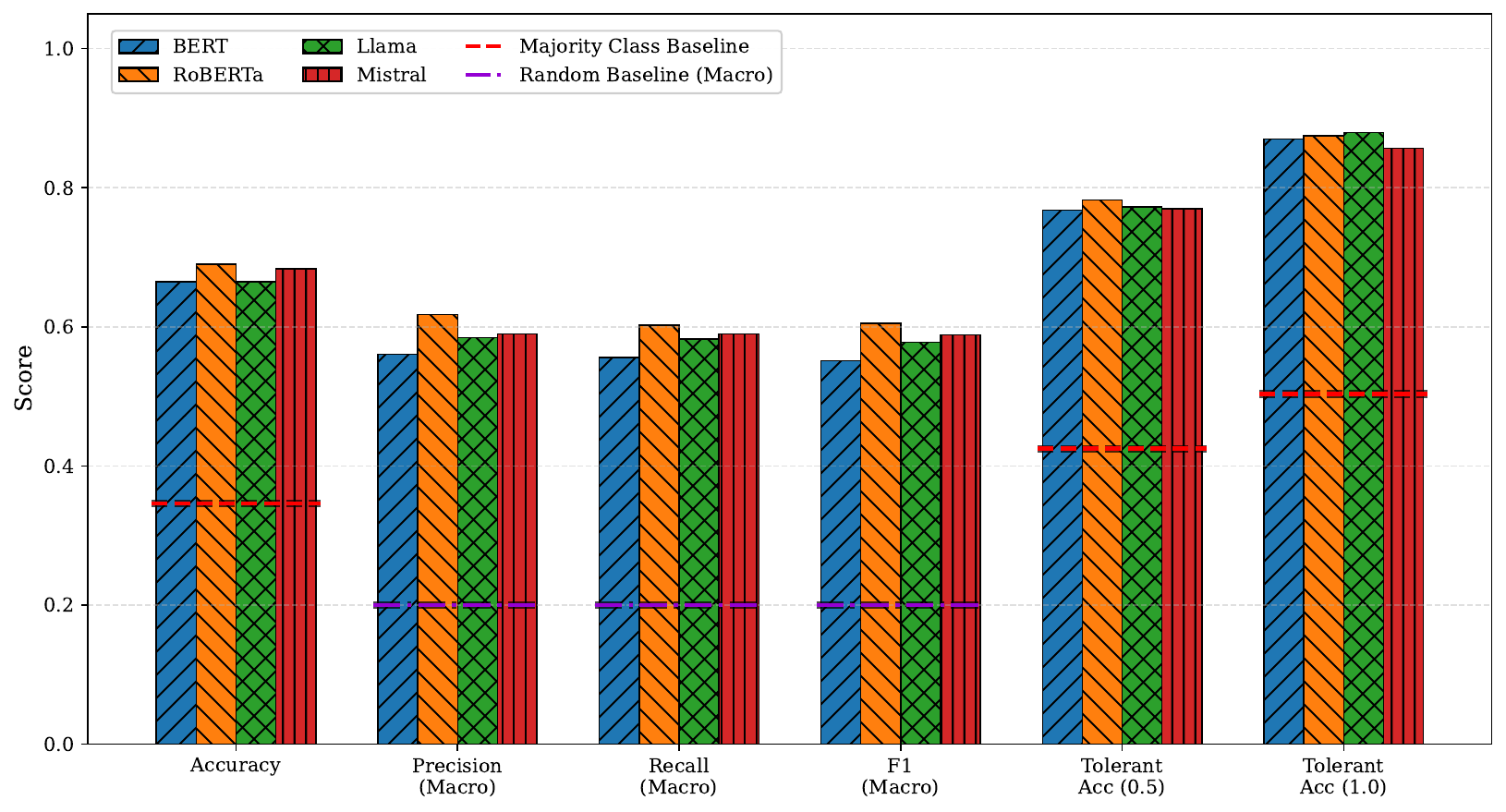}
  \caption{Publisher reliability prediction: performance metrics for BERT, RoBERTa, Llama, and Mistral. Dashed red lines indicate the Majority Class Baseline (for Accuracy and Tolerant Accuracy); dash-dotted violet lines indicate the Random Baseline at 0.20 (for Macro metrics).}
  \label{fig:metrics}
\end{figure}

Figure \ref{fig:metrics} presents the results. The bar chart benchmarks performance against two distinct baselines, represented by dashed lines:
\begin{itemize}
    \item \textbf{Majority Class Baseline (MCB) (Red):} Represents a model that always predicts the most frequent domain label. It serves as the reference for Accuracy and Tolerant metrics. Note that the three red lines differ: for example, a majority predictor achieves a given Accuracy by always predicting $l_4$, but under Tolerant Accuracy (0.5) it also receives half credit for $l_3$ and $l_5$ predictions that fall one step away, and full credit under Tolerant Accuracy (1.0), progressively raising the baseline value across the three metrics.
    \item \textbf{Random Baseline (RB) (Violet, 0.20):} Represents a uniform random guess ($1/N$). This is the reference for Macro metrics, as they weight all classes equally regardless of their size.

\end{itemize}

With an Accuracy of 0.69 RoBERTa leads, followed closely by Mistral at 0.68. These scores are more than triple those of the Random Baseline. 
Considering $\mathrm{Accuracy}_{\mathrm{tolerant}}$  (1.0), Llama achieves the top score of 0.88, followed by RoBERTa (0.87) and Mistral (0.86). These scores substantially outperform the majority class baseline.

Publisher-level errors are strictly local and concentrate between adjacent reliability levels, most notably between $l_4$ (Generally Credible) and $l_5$ (High Credibility); the Supplementary Material~\cite{bianchi2026supplementary} reports full publisher-level confusion matrices and a detailed error analysis.

\section{Discussion and Conclusion}
\label{sec:conclusions}
In this study, we investigated whether the reliability of news publishers could be inferred from their articles. We framed the task as a five-level ordinal classification problem aligned with NewsGuard ratings and evaluated four transformer-based models, BERT, RoBERTa, Llama, and Mistral, under a strict publisher-disjoint protocol to simulate real-world deployment on unseen sources.

Our findings demonstrate that language models can capture meaningful and generalizable signals of publisher reliability. Interestingly, the substantially larger Mistral 24B does not outperform smaller architectures, suggesting that performance is limited by dataset characteristics rather than model size. Despite operating with a 1,024-token context window (four times larger than the 256-token limit used for the other models), Mistral shows no measurable gain in accuracy or F1 score. This may indicate that the cues discriminating reliable publishers are concentrated in the beginning of articles. The lack of benefit from the extended context further supports the hypothesis that increasing model size or input length alone cannot overcome limitations imposed by weak supervision. In weak supervision, each article inherits the reliability label of its publisher. Therefore, an individual article's deviation from the outlet's overall NewsGuard rating introduces label noise. Additionally, the diversity of publisher-level signals is limited. The moderate article-level accuracy should be interpreted in light of the ultimate objective of the framework. Since publisher reliability is a source-level property, individual articles provide only noisy evidence. The substantial performance gain obtained after aggregation suggests that reliability emerges more clearly when multiple articles are considered jointly. A key finding of this study is that publisher reliability should not be modeled as a property of isolated articles. Rather, it emerges as a collective signal that becomes substantially more stable and informative when evidence is aggregated across multiple articles from the same source. This observation supports a shift from article-centric formulations toward source-level inference frameworks, where article classification serves as an intermediate step rather than the final objective.


At the same time, our bias analysis reveals systematic interactions between reliability prediction and political leaning. Models tend to overestimate the reliability of politically extreme outlets and underestimate slightly left-leaning ones. While effect sizes remain moderate, these patterns highlight the need for bias-aware evaluation protocols in automated reliability assessment systems.

The study also presents important scope constraints. First, the dataset is restricted to English-language political news outlets, limiting generalizability to other linguistic and thematic ecosystems. Second, the weak supervision setting, where article labels inherit publisher-level ratings, introduces structural noise, as individual articles may deviate from their outlet's overall reliability profile. Finally, publisher-level inference relies on simple majority voting, which does not yet incorporate prediction confidence or uncertainty. 

Future research should extend this framework to multilingual settings, explore confidence-aware aggregation strategies, and develop training objectives explicitly tailored to ordinal classification. More broadly, combining content-based approaches with ecosystem-level signals, such as ownership transparency, citation networks, or audience interactions, may provide a more comprehensive and robust framework for scalable news source reliability assessment.

\bibliographystyle{IEEEtran}
\bibliography{bibliography}

\end{document}